# Authorship clustering using multi-headed recurrent neural networks

## Notebook for PAN at CLEF 2016

Douglas Bagnall

douglas@halo.gen.nz

**Abstract.** A recurrent neural network that has been trained to separately model the language of several documents by unknown authors is used to measure similarity between the documents. It is able to find clues of common authorship even when the documents are very short and about disparate topics. While it is easy to make statistically significant predictions regarding authorship, it is difficult to group documents into definite clusters with high accuracy.

## 1 Introduction

The most successful entry in the PAN 2015 author identification task [7][2] used a form of recurrent neural network (RNN) to simultaneously model the language of several authors. The relative success of each author's model when presented with an anonymous text was treated as an indication of true authorship. This technique is reused here, but with different interpretive steps to suit the different task.

The use of recurrent neural networks for language models is not new, and was most recently revived by Mikolov [6]. The novelty of [2] was the use of a single recurrent state that was shared by multiple language models, reducing both overfitting and computational cost. The system produces scores in the form of relative entropies which suit the attribution problem well because it avoids the problems of high dimensional feature space, cutting directly to pairwise similarity scores.

While this suited the PAN2015 author identification task[7], and the combination of language modelling and information theory is clearly useful in uncovering authorship, the approach may have drawbacks when used for clustering. At its core it produces an asymmetrical matrix of pair-wise divergence scores. Lacking both symmetry and the triangle inequality, this matrix cannot be used with clustering algorithms designed for metric spaces — which is to say most of them.

This paper briefly describes the multi-headed recurrent neural network introduced in [2], then looks at a method of turning its output into clustering decisions. But first comes a description of the task, noting in particular the scoring mechanisms which will come to contort the rest of the work.

### 1.1 The PAN 2016 author clustering task

For a full description of the competition, see the overview paper [7]. The following concise definition is taken from the PAN website: [1]

---

[1] http://pan.webis.de/clef16/pan16-web/author-identification.html

Given a collection of (up to 100) documents, identify authorship links and groups of documents by the same author. All documents are single-authored, in the same language, and belong to the same genre. However, the topic or text-length of documents may vary. The number of distinct authors whose documents are included in the collection is not given.

The task covers three alphabetic languages (English, Greek, and Dutch), with six problems in each language. As described in the quoted passage, each problem consists of up to 100 documents. Two forms of answer are required for each problem: a set of clusters, indicating texts presumed to be by a single author; and a set of weighted links between text pairs where a higher weight relates to a higher probability that the two texts are from the same author. These two outputs are scored in different ways. The clustering are evaluated using the *F(BCubed)*[1] measure which averages the precision and recall of each document. The weighted links are scored using *mean average precision*[5] (or *MAP*), which punishes for every false link that is scored higher than a true link. Although the scores are presented in the form of probabilities, MAP doesn't actually care about their relative values, just their rank.

## 2  The multi-headed recurrent neural network language model

This description is simplified for brevity; for more detail see [2] or [6] for an overview of RNN based language modelling.

A standard character-level language model will, given an unfinished sequence of text $x_1, x_2, x_3, \ldots, x_{i-1}$, predict the next character $x_i$. That is, it outputs a probability distribution over the possible characters $p(x_i|x_{i-1}, x_{i-2}, \ldots, x_1)$. Such a model is usually trained on a corpus of text and the probabilities it emits are based on that text (and of course its structure and meta-parameters). Given the hypothesis that the writing style of an author will inevitably be reflected in choices made at the character level (even if that relationship is very faint), it follows that a language model trained solely on one author's work is likely to better predict another text by that author than would a model trained on the work of another author.

An RNN based language model will usually have a *softmax* activation for the output layer $z$. Where there are $k$ output nodes (corresponding to the set of possible symbols), softmax for node $j$ is defined as

$$\sigma(\mathbf{z})_j = \frac{e^{z_j}}{\sum_k e^{z_k}}$$

which provides values that can be treated as mutually exclusive probabilities. The multi-headed RNN language model differs in that it simultaneously models the language of many documents at once by using multiple softmax groups. Given $M$ documents, there are $M * k$ output nodes arranged in $M$ independent softmax groups. Each of these groups is trained primarily on a single text, with some stochastic "leakage" from other texts which helps regulate the output layer weights, preventing gross overfitting. The error gradient is back-propagated to the shared hidden layer.

In most regards the network follows the basic structure described by Elman [4], which is to say it resembles a multi-layer perceptron with a single hidden layer modified so the hidden state depends in part on the previous iteration's hidden state.

At each time step $t$, the hidden state $h_t$ depends on the hidden state at the previous time step $h_{t-1}$ as well as the input vector $x_t$ which represents a single character. Where $b_h$ is a bias vector, $W_{xh}$ and $W_{hh}$ are weight matrices, and $f_h$ is a non-linear function, the update of the hidden state is $h_t = f_h(W_{hh}h_{t-1} + W_{xh}x_t + b_h)$. An output vector $y_t$ is derived from the hidden state, with $y_t = f_y(W_{hy}h_t + b_y)$. For this work, as in [2], the "ReSQRT" function is used for $f_h$:

$$f_h(x) = \begin{cases} \sqrt{x+1} - 1 & \text{if } x \geq 0 \\ 0 & \text{otherwise.} \end{cases}$$

The input layer uses a "one-hot" representation of the symbols; there is a node for each of the $N$ symbols in the alphabet, and at each time step the node for the current symbol is set to 1 while the rest are 0.

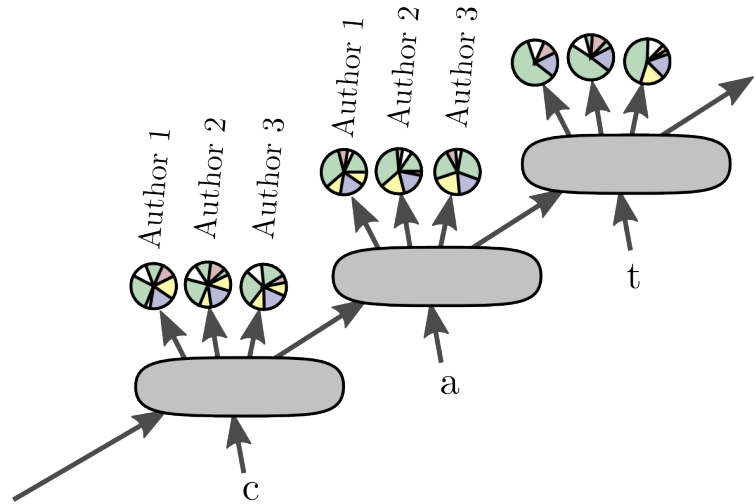

**Fig. 1.** The multi-headed RNN language model has multiple sub-models, each making predictions based primarily on a single unique text. If some aspects of the text's author's style is captured by the sub-model, other texts by the same author are likely to be relatively well predicted by it.

The network is trained using a variant of adagrad[3] and back-propagation through time (BPTT). Simply described this involves iteratively adjusting the weight matrices with an individual monotonically decreasing learning rate for each weight.

The recurrent hidden layer can be thought of as modelling the language as a whole, while the various sub-models pick out aspects of the recurrent state that suit their document.

## 3  Method

### 3.1  Text preprocessing

In order to simplify the computational task and remove the distorting effect of extremely rare characters, all the texts were mapped to a smaller character set following the method described in [2]. The following description is an abbreviated version of a section of that paper. In addition, in two out of the five runs for each language, rare words were replaced by special tokens (Section 3.2).

The text is converted into the NFKD unicode normal form, which decomposes accented letters into the letter followed by the combining accent. Capital letters are further decomposed into an uppercase marker followed by the corresponding lowercase letter.

Various rare characters that seem largely equivalent are mapped together; for example the en-dash ("–") and em-dash ("—") are rare and appear to be used interchangeably in practice so these are mapped together.

For the Greek text, all Latin characters are mapped to a single token (the letter `s`) on the basis that foreign quotations and references appear too rarely for their content to be valuable and an attempt to model them would be wasteful, but the tendency to use them might be a useful signal. Following similar logic, all digits in all languages are mapped to 7. Runs of whitespace are collapsed into a single space.

At the end of this processing, any character with a frequency lower than 1 in 10,000 is discarded. Any characters occurring in a text but not in the resultant alphabet are ignored—there is no "unknown" token. Alphabet sizes are 45 for English, 47 for Dutch, and 51 for Greek.

Runs of more than 5 identical characters are truncated at 5. This is mainly aimed at the Latin stretches in Greek text, where the exact word length is probably not a useful signal.

### 3.2  Eliminating low document frequency words

There are always going to be character level patterns in text that are more indicative of topic or genre than authorial style. Genre is to ostensibly controlled in the PAN corpora, but topic is not. A content-agnostic language model trained on a text containing topic-specific words will expect those words to appear frequently.

For example, this is the entirety of a text used in the Dutch review training problems:

```
Ik heb de Blackberry 9790 Bold nu sinds vijf dagen, en ik
ben zeer tevreden over deze telefoon. Ik was op zoek naar
een telefoon zonder te veel poespas en onzinnige applicaties,
maar die wil snel en gemakkelijk werkt. Het menu werkt snel en
instinctief. Als ik iets wil veranderen, heb ik het zo gevonden.
Tijdens vergaderingen gebruik ik soms de luidspreker; dat werkt
perfect. Ik heb overigens verschillende recensies gelezen over
een slechte batterij, maar mijn blackberry houdt het wel de
hele dag vol. Volgens mij hebben andere smartphones veelal
dezelfde problemen.
```

The word `blackberry` doesn't occur anywhere else in the corpus, yet it accounts for about 3.5% of this text. A naively trained language model could be forgiven for assuming that a propensity to write `blackberry` is a trademark of this author, which is unlikely in reality.

To counter this for some runs words that occur in only a very few documents (including in the controls) are replaced by a rare word token (arbitrarily, the degree sign °). The following quote is the above review modified according to the rules described with the words occuring in fewer than 1 percent of documents replaced with ° tokens:

```
¹ik heb de ¹° 7777 ¹° nu sinds vijf dagen, en ik ben zeer
tevreden over deze telefoon. ¹ik was op zoek naar een telefoon
zonder te veel ° en ° °, maar die wil snel en gemakkelijk werkt.
¹het menu werkt snel en °. ¹als ik iets wil veranderen, heb ik
het zo gevonden. ¹tijdens vergaderingen gebruik ik soms de °;
dat werkt perfect. ¹ik heb overigens verschillende ° gelezen
over een slechte batterij, maar mijn ° houdt het wel de hele
dag vol. ¹volgens mij hebben andere smartphones ° dezelfde
problemen.
```

While this clearly removes some topic specific words (`Blackberry`, `Bold`, `applicaties`, `luidspreker`), it also mangles some possibly useful and seemingly ordinary words and phrases (`poespas  en  onzinnige`, `veelal`).[2] It is difficult to ascertain whether this is of net benefit on the small training set, or indeed what the threshold should be. For each language an ensemble of five models was used; for two of these the document frequency threshold was used. There are around 300 documents for each language (see next section) so the lowest effective threshold is in the order of 0.005, corresponding to the word occurring in a single document.

It would likely be of benefit to take part-of-speech information into account when discarding words, but that would complicate things and involve tagging software that is not unavailable for all languages.

### 3.3  MHRNN set up

Although in both the training and test sets there are six independent problems for each language, there is (at least in the training set) overlap between the sets with some texts being in multiple problems. As a result, although the training problem sizes sum to 390, 471, and 330 texts for English, Dutch, and Greek respectively, there are only 201, 278, and 189 individual texts, also respectively. Numbers for the tests sets are unknown.

All the documents from all the problems in each language were combined into a single model, along with 80 "control" texts selected at random from the 2013, 2014, and 2015 PAN training corpora. That means that in the training set for the Greek model there were 269 (i.e. 189 problem texts + 80 control) softmax groups; for the Dutch 358, and for the English 281.

Calculating all the problems at once is beneficial in a number of ways. The more text that the model as a whole sees, the better it can model the target language. More text allows the hidden layer to be larger, giving it more subtlety. Treating each problem individually with a larger portion of control texts (e.g. 250 control texts per problem) would possibly work, at great computational cost. On the other hand, the presence of the other problems may be beneficial as they teach the model more about the genres in question. Many of the control texts in English (for example) are excerpts from bad 19th century novels, which differ significantly from the PAN2016 problem texts.

---

[2] I do not speak Dutch and can only guess at the value of these words.

### 3.4 Interpreting the results

The cross-entropy of each sub-model running against each text is collected in a matrix. For every problem the relevant sub-set of the matrix (i.e. problem texts evaluated by problem models) is gathered up. Because some texts are inherently more difficult to model than others, the problem matrix is adjusted by subtracting the mean score given to each text by the control models. This gives the problem matrix a mean of approximately zero.

The control models were used for this (rather than the models of the problem itself, or those of other problems) because between them the problems share authors across many texts. Using these models to normalise scores might skew the matrix if a few authors dominate. On the other hand, some of the control texts are known to possess peculiar styles while others could well be by the same authors as the the problem texts (given that they are from previous competitions and PAN draws corpora from limited pools). No attempt was made to find out.

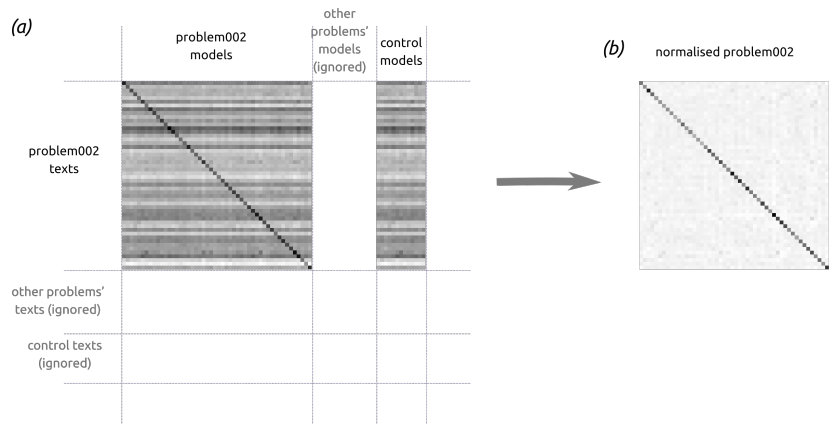

**Fig. 2.** Extracting normalised relative entropies for a problem (in this case, problem002 in the training set). The control models are used to normalise the text scores. Darker colours mean more affinity; the diagonal shows that each model is good at modelling the text it was trained on. Sub-figure b has been made symmetric and positive (see text).

The matrix is then added to its own transpose for symmetry and made positive and monotonically increasing by exponentiation. That is given the normalised matrix $M$ is converted thus $M' = e^{M+M^T}$. The top triangle of $M'$ is scaled to the range 0–1 for the MAP scores. This simple strategy seems to work reasonably well for MAP scores.

### 3.5 Optimising F(BCubed): the cowardly approach

By definition, the F(BCubed) metric must be above $0.5$ when all documents are placed in their own individual clusters of size one (because this makes the precision of each document 1, while recall is at worst $1/N$ when all documents belong to the same cluster). On the other hand, placing all documents in a single big cluster will in result in an F-score less than $0.5$ in the *typical* case.[3] This reflects that the fully disconnected solution states only the a-priori truth that each document is in a cluster with itself, and F(BCubed) rewards the restraint of that claim.

Therefore, given no other information, the optimal strategy is predict $N$ fully disconnected clusters of size 1. It only makes sense to depart from this strategy when the underlying detection is strong. In this paper, the fully disconnected solution is called the *cowardly strategy*, and the rest of this section is devoted to detecting ways in which it might be bettered.

Figure 3 shows the problem.

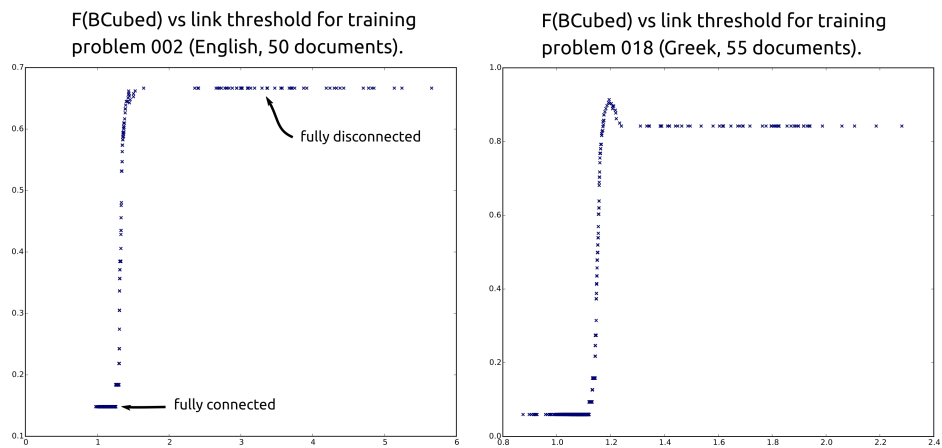

**Fig. 3.** The x axis is the threshold at which to link two texts, using simple agglomerative clustering. The y axis is the resultant F(BCubed) score. In the bottom left, all the documents are in one cluster and all the links are redundant. The flat line at the top right corresponds to the diagonal of the affinity matrix: these are the thresholds at which documents would link to themselves. In between is a large cliff. Sometimes there is a little hill at the top of the cliff. The task is to climb the hill without falling off the cliff. It is completely dark. You don't know where the cliff is, or whether there is a hill. Where there is no hill (as with problem002), you are better off sticking to the right. That is the *cowardly* strategy which this paper struggles to better. The exact values on both axes are largely irrelevant.

### 3.6 Optimising F(BCubed): the cluster-aware approach

---

[3] Typical not only in the sense of a randomly sampled partition, but more importantly this is empirically true for all the training set problems.

One problem with the simple agglomerative approach is that the a link between two documents can cluster together a large number of other documents that might otherwise seem unrelated. This is equivalent to single-linkage in the metric clustering case. Figure 4 attempts to illustrate the problem. As clusters get big, the probability of a single link leading to cataclysmic super-cluster grows, causing the cliff in Figure 3.

A modified agglomerative approach was developed where each link's score is adjusted to the mean of all the links in the cluster it forms. This approach, which appeared to give better results, is not described here because it was mistakenly not used in the PAN evaluation.

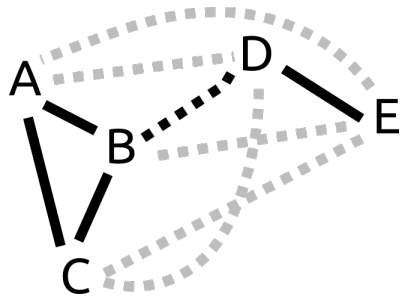

**Fig. 4.** Linking $B$ to $D$ seems attractive, but it implicitly forms the links $A - D$, $A - E$, $B - E$, $C - D$, and $C - E$; and the quality of those links needs to be taken into account. With the cluster-aware approach, the link is scored with the mean of all the links in the resulting cluster.

### 3.7 Optimising F(BCubed): the accidental small-cluster steep cliff approach

Due to a foolish programming error, the cluster-aware strategy was replaced by an algorithm was used that effectively punished any link that joined more than two documents together. That is, the documents were all made to partner up before any of them could consider larger clusters.

Although this error seems drastic, the algorithm turns out to have some nice features. Figure 5 shows the modified F(BCubed) curves for the same examples as Figure 3. The cliff is steeper and the hill, where there is one, is broader and flatter (though lower). This makes aiming at better-than-cowardly easier using the simple heuristic described in the next section. Results on the training set using the same heuristic and the intended algorithm (as in Section 3.6) are in aggregate very similar to those obtained with this accidental method, though the variance is larger. The intended algorithm appears to make higher scores achievable in a narrower band of thresholds; outside the band the scores are worse.

### 3.8 Optimising F(BCubed): the clusteriness heuristic

The aim of these clustering explorations is to find a method that beats the cowardly strategy. This appears achievable by using the simple heuristic of finding anchor points in the F(BCubed) landscape, as shown in Figure 5, and choosing a point between them according to a fixed ratio. The exact ratio was chosen per language and genre based on a terrible mixture of greed and fear. In the end a slightly risky coefficient was chosen as the underlying detection seems quite sound (reflected in significantly better than random MAP scores on the training set), and if sticking to the cowardly baseline is the best strategy it is almost certainly going to result in a tie. Thus safety is discarded in pursuit of a win.

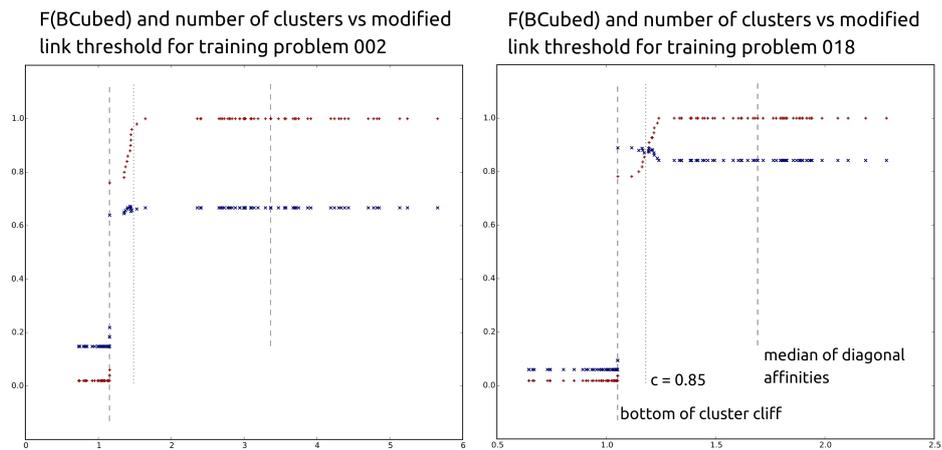

**Fig. 5.** The x axis is the threshold at which to link two texts, using an odd algorithm that prefers documents to form monogamous pairs rather than large clusters. F(BCubed) is shown with blue ×, and the number of clusters is shown with red +, and has been scaled to fit the same y axis (so 1 means $N$ clusters, $1/N$ means 1 cluster). The number of clusters is known to the clustering algorithm while the F(BCubed) score is not. The algorithm uses the bottom of the "cluster cliff" and the median of the diagonal values (that is, the love models have for their own documents) as anchors, marked as grey dotted line. A threshold is chosen between these points using a predetermined *clusteriness* coefficient (in this case is $0.85$), simply by reaching that far from the diagonal anchor to the cliff anchor. Calling the clusteriness $c$, the cliff and diagonal anchors $t_c$ and $t_d$, and the final threshold $t$ yields $t = t_d - c * (t_d - t_c)$. This clusteriness heuristic also works with the simple agglomerative or cluster aware strategies, though the cliff is not so steep and the landscape at the top is more exaggerated.

## 3.9 Optimising F(BCubed): example training set results

| Lang/genre | problem | MAP | coward | best | $c_b$ | diff | fixed | $c_f$ | diff |
|---|---|---|---|---|---|---|---|---|---|
| en articles | problem001 | 0.028 | 0.824 | 0.824 | 0.80 | 0.000 | 0.817 | 0.82 | -0.007 |
| en articles | problem002 | 0.110 | 0.667 | 0.667 | 0.79 | 0.000 | 0.662 | 0.82 | -0.005 |
| en articles | problem003 | 0.054 | 0.925 | 0.925 | 0.87 | 0.000 | 0.925 | 0.82 | 0.000 |
| en reviews | problem004 | 0.051 | 0.815 | 0.815 | 0.77 | 0.000 | 0.811 | 0.79 | -0.004 |
| en reviews | problem005 | 0.018 | 0.933 | 0.933 | 0.83 | 0.000 | 0.933 | 0.79 | 0.000 |
| en reviews | problem006 | 0.114 | 0.667 | 0.707 | 0.90 | **0.040** | 0.664 | 0.79 | -0.003 |
| nl articles | problem007 | 0.407 | 0.944 | 0.955 | 0.91 | **0.011** | 0.944 | 0.81 | 0.000 |
| nl articles | problem008 | 0.380 | 0.659 | 0.782 | 0.94 | **0.123** | 0.675 | 0.81 | **0.017** |
| nl articles | problem009 | 0.128 | 0.825 | 0.845 | 0.86 | **0.020** | 0.833 | 0.81 | **0.008** |
| nl reviews | problem010 | 0.122 | 0.701 | 0.726 | 0.77 | **0.025** | 0.723 | 0.77 | **0.021** |
| nl reviews | problem011 | 0.039 | 0.802 | 0.806 | 0.71 | **0.004** | 0.799 | 0.77 | -0.003 |
| nl reviews | problem012 | 0.002 | 0.953 | 0.953 | 0.78 | 0.000 | 0.953 | 0.77 | 0.000 |
| gr articles | problem013 | 0.282 | 0.675 | 0.742 | 0.95 | **0.068** | 0.675 | 0.85 | 0.000 |
| gr articles | problem014 | 0.218 | 0.817 | 0.862 | 0.92 | **0.045** | 0.826 | 0.85 | **0.008** |
| gr articles | problem015 | 0.236 | 0.932 | 0.939 | 0.87 | **0.007** | 0.939 | 0.85 | **0.007** |
| gr reviews | problem016 | 0.270 | 0.952 | 0.962 | 0.81 | **0.010** | 0.954 | 0.82 | **0.001** |
| gr reviews | problem017 | 0.462 | 0.675 | 0.817 | 0.88 | **0.142** | 0.735 | 0.82 | **0.061** |
| gr reviews | problem018 | 0.585 | 0.842 | 0.914 | 0.83 | **0.072** | 0.903 | 0.82 | **0.061** |

**Table 1.** Indicative MAP and F(BCubed) scores derived from one run on the training set. The *coward* column shows the F(BCubed) score for a fully disconnected solution where each document is in its own cluster. The *best* column shows the maximum score obtainable using the techniques described here, while the next column ($c_b$) shows the clusteriness setting necessary to obtain that score. The *fixed* column show the scores obtained by setting the clusteriness according to heuristics based on language and genre, and the corresponding $c_f$ column shows that setting. The two *diff* columns show the difference between the cowardly approach and the *best* and *fixed* scores respectively; gains are highlighted.

Table 1 shows some results obtained on the training set. It shows that for most of the English problems there is no chance of getting a better-than-cowardly result (using these techniques), but for the other languages this is not only possible but is achieved using the preset values. Those preset values were of course adjusted with full knowledge of the training set.

### 3.10 Ensembles

Five nets were trained for each language, and the raw cross entropy matrices were summed before subsequent processing. All the models were trained in a similar fashion, using a form of adagrad optimisation, with somewhat haphazard variation in a few meta-

parameters. A summary of the meta-parameters is shown in Table 2.[4]

Whether this variation in meta-parameters (or indeed the use of ensembles at all) actually helps was not thoroughly explored.

|  | size | PSN | leak | over-fit | direction | word DF |
|---|---|---|---|---|---|---|
| Dutch | 299 | 0.5 | $\frac{1}{2N}$ | 4 | forward | - |
|  | 159 | 0.3 | $\frac{1}{3N}$ | 2 | forward | - |
|  | 139 | 0.3 | $\frac{1}{2N}$ | 4 | *reverse* | 0.005 |
|  | 99 | 0.5 | $\frac{1}{2N}$ | 3 | forward | 0.01 |
|  | 139 | 0.3 | $\frac{1}{2N}$ | 5 | *reverse* | - |
| English | 299 | 0.5 | $\frac{1}{2N}$ | 4 | forward | - |
|  | 139 | 0.3 | $\frac{1}{2N}$ | 5 | *reverse* | - |
|  | 239 | 1 | $\frac{1}{3N}$ | 2 | *reverse* | 0.005 |
|  | 139 | 0.3 | $\frac{1}{2N}$ | 5 | forward | 0.01 |
|  | 159 | 0.5 | $\frac{1}{2N}$ | 2 | forward | - |
| Greek | 299 | 0.3 | $\frac{1}{2N}$ | 3 | forward | 0.005 |
|  | 279 | 0.5 | $\frac{1}{2N}$ | 4 | forward | - |
|  | 159 | 0.3 | $\frac{1}{3N}$ | 5 | *reverse* | - |
|  | 159 | 1 | $\frac{1}{2N}$ | 3 | forward | 0.005 |
|  | 139 | 0.3 | $\frac{1}{2N}$ | 5 | *reverse* | - |

**Table 2.** Meta-parameters that varied across training runs. *size* is the number of neurons in the network's hidden layer (for slight efficiency reasons, one less than a multiple of four is always chosen). *PSN* is the standard deviation of Gaussian pre-synaptic noise added during training. *leak* is the chance that each training example has of affecting a non-target softmax group in the first epoch (subsequently it decays exponentially). *Over-fit* is the number of epochs that training continues after the average cross-entropy against a validation text has started worsening. *word DF* refers to the document frequency threshold for word inclusion; a dash means no words are removed.

As the number of documents is much greater than typically found in the PAN2015 challenge, the hidden layers can be larger than seen in [2], resulting in hopefully more accurate and nuanced modelling at the cost of training time.

In ordinary language modelling, the point is to achieve maximum accuracy, and to this end a validation corpus is often used to avoid overfitting the model to the training set. For this task a similar approach is used, though accuracy of the language model itself is of course not the primary concern. It was found that a slightly overfit model (vis-a-vis a validation text, averaged across all sub-models) seems to give better MAP results. Hence the nets were trained until the validation entropy had been worsening for a small number of epochs.

---

[4] Full meta-parameter details are defined in code at https://github.com/douglasbagnall/bog/tree/master/config; there is little to be gained from an exhaustive summary.

For each language, two of the five models read the texts backwards, learning to predict the characters that lead to the current state. While reversed language models are no better than forward ones on their own, they were included on the hypothesis that their differing perspective should help the ensemble.

There were also two nets for each language with a word document frequency threshold or $0.005$ or $0.01$. The arrangement of all the meta-parameters is essentially ad-hoc.

## 4 Results

| Lang | genre | problem | F(BCubed) | R-BCubed | P-BCubed | MAP |
|---|---|---|---|---|---|---|
| en | articles | problem001 | 0.83333 | 0.71429 | 1 | 0.11628 |
| | | problem002 | 0.66667 | 0.5 | 1 | 0.36283 |
| | | problem003 | 0.95522 | 0.91429 | 1 | 0.040784 |
| | reviews | problem004 | 0.84058 | 0.725 | 1 | 0.096951 |
| | | problem005 | 0.94172 | 0.9 | 0.9875 | 0.028065 |
| | | problem006 | 0.6825 | 0.525 | 0.975 | 0.10814 |
| nl | articles | problem007 | 0.84121 | 0.74561 | 0.96491 | 0.26479 |
| | | problem008 | 0.91896 | 0.87719 | 0.96491 | 0.05726 |
| | | problem009 | 0.68543 | 0.52632 | 0.98246 | 0.18788 |
| | reviews | problem010 | 0.9418 | 0.89 | 1 | 0.071431 |
| | | problem011 | 0.67947 | 0.52 | 0.98 | 0.054028 |
| | | problem012 | 0.82343 | 0.71 | 0.98 | 0.018187 |
| gr | articles | problem013 | 0.84298 | 0.72857 | 1 | 0.24491 |
| | | problem014 | 0.66667 | 0.5 | 1 | 0.28022 |
| | | problem015 | 0.93939 | 0.88571 | 1 | 0.014714 |
| | reviews | problem016 | 0.86022 | 0.82381 | 0.9 | 0.3739 |
| | | problem017 | 0.89143 | 0.92857 | 0.85714 | 0.17308 |
| | | problem018 | 0.79028 | 0.65952 | 0.98571 | 0.54633 |

**Table 3.** Results for this paper in the 2016 PAN Author Clustering evaluation. *P-BCubed* and *R-BCubed* are the precision and recall constituent parts of the *F(BCubed)* score. *MAP* is mean average precision.

Results are shown in Table 3. At the time of writing it is unknown how these results compare to other approaches.

Where the *P-BCubed* column in Table 3 is 1, it is likely (but not certain) that the software has settled on the cowardly strategy (Section 3.5). Where it is not 1, the software has deviated from this strategy, whether successfully or not. There is cause for optimism where the MAP scores are greater than $0.1$ and the BCubed precision is not far below $1$. On the whole the software looks to have favoured valour over caution.

## 5 Discussion

The technique described *seems* to perform reasonably well on an intrinsically difficult problem, but it is sadly difficult to be certain how it compares to other methods. Most entries in the PAN competition appear to have suffered problems relating to their understanding of the scoring system, which may be hiding some successes at actually detecting links between texts.

### 5.1 A zero-effort baseline

| Lang | genre | problem | MAP | random MAP |
|---|---|---|---|---|
| en | articles | problem001 | 0.028 | 0.043 |
| en | articles | problem002 | 0.110 | 0.132 |
| en | articles | problem003 | 0.054 | 0.060 |
| en | reviews | problem004 | 0.051 | 0.014 |
| en | reviews | problem005 | 0.018 | 0.004 |
| en | reviews | problem006 | 0.114 | 0.024 |
| nl | articles | problem007 | 0.407 | 0.004 |
| nl | articles | problem008 | 0.380 | 0.059 |
| nl | articles | problem009 | 0.128 | 0.016 |
| nl | reviews | problem010 | 0.122 | 0.017 |
| nl | reviews | problem011 | 0.039 | 0.010 |
| nl | reviews | problem012 | 0.002 | 0.006 |
| gr | articles | problem013 | 0.282 | 0.029 |
| gr | articles | problem014 | 0.218 | 0.017 |
| gr | articles | problem015 | 0.236 | 0.010 |
| gr | reviews | problem016 | 0.270 | 0.029 |
| gr | reviews | problem017 | 0.462 | 0.065 |
| gr | reviews | problem018 | 0.585 | 0.103 |

**Table 4.** Comparing actual training set MAP scores from this paper with MAP scores from randomly shuffled score matrices. The randomised values are from a single run; repetitions of the experiment will give different results.

The PAN committee published a baseline result on an early-bird test set with an F(BCubed) score of $0.6922$, and a MAP of $0.000459$. As has been well established, the fully-disconnected cowardly strategy offers a hard-to-beat, easy-to-achieve baseline for F(BCubed). For MAP a beatable but useful baseline might be a fully connected graph with random link strengths. Randomly shuffling the link arrays gives results like the final column in Table 4. The average of this column is $0.036$ – two orders of magnitude better than the "official" baseline. MAP rewards verbosity. Even if a method gives no ranking of undesired links, it is better to assign them low random weights than to ignore them altogether.

As explained at length in Section 3.5, a system that sticks to the a-priori truth that each document is in a cluster with itself will obtain an F(BCubed) of around $0.8$. Thus it is easy to define a zero-effort baseline that randomizes links for MAP and uses the cowardly strategy for F(BCubed). Unfortunately this baseline would have performed quite well in competition. Many teams seem not to have grasped the fundamental biases of the two scoring mechanisms, although their underlying solutions may be sound.

This suggests a weakness in F(BCubed)[1] for evaluating very difficult clustering problems. Whereas from an informational point of view putting everything in a single cluster might not seem very different from putting everything in separate clusters of size one, F(BCubed) over-rewards the latter claim. It may be that there is no clustering evaluation system that suits all problems.

MAP and F(BCubed) are evidently both quite tricky to code and reason about. It might help in future competitions if some form of evaluation software was available so that inexperienced coders could gain a better understanding of the task and their progress in it.

### 5.2 Variations and run-time

A major drawback to this technique is the time it takes. As described in the paper it took far longer than any other technique. An obvious way to speed up the process would be to reduce the ensemble size to one (for a five-fold improvement). Reducing the number of hidden neurons would further improve speed but reduce efficacy a small margin.

Going the other way, increasing the amount and quality of control text would increase the system's overall understanding of the language, and allow the number of hidden nodes to be increased. This should lead to somewhat better results without fundamental changes.